\documentclass[11pt,a4paper]{article}
\usepackage[hyperref]{emnlp-ijcnlp-2019}
\usepackage{times}
\usepackage{latexsym}
\usepackage{graphicx}

\usepackage{url}
\usepackage{booktabs}
\usepackage{multirow}
\usepackage{amsmath}
\usepackage{amssymb}

\DeclareMathOperator*{\argmin}{argmin}

\aclfinalcopy 

\title{Pivot-based Transfer Learning for Neural Machine Translation\\between Non-English Languages}

\author{Yunsu Kim$^{1\hspace{-0.1em}}\Thanks{\hspace{0.5em}Equal contribution.}$ \hspace{6pt} Petre Petrov$^{1,2*}$ \hspace{4pt} Pavel Petrushkov$^{2}$ \hspace{6pt} Shahram Khadivi$^{2}$ \hspace{6pt} Hermann Ney$^{1}$\\
  $^{1}$RWTH Aachen University, Aachen, Germany\\
  {\tt \{surname\}@cs.rwth-aachen.de}\\
  $^{2}$eBay, Inc., Aachen, Germany\\
  {\tt \{petrpetrov,ppetrushkov,skhadivi\}@ebay.com}\\}

\date{}

\begin{document}
\maketitle
\begin{abstract}
  We present effective pre-training strategies for neural machine translation (NMT) using parallel corpora involving a pivot language, i.e., source-pivot and pivot-target, leading to a significant improvement in source$\rightarrow$target translation.
  We propose three methods to increase the relation among source, pivot, and target languages in the pre-training: 1) step-wise training of a single model for different language pairs, 2) additional adapter component to smoothly connect pre-trained encoder and decoder, and 3) cross-lingual encoder training via autoencoding of the pivot language.
  Our methods greatly outperform multilingual models up to +2.6\% \textsc{Bleu} in WMT 2019 French$\rightarrow$German and German$\rightarrow$Czech tasks.
  We show that our improvements are valid also in zero-shot/zero-resource scenarios.\\
\end{abstract}

\section{Introduction}
  Machine translation (MT) research is biased towards language pairs including English due to the ease of collecting parallel corpora.
  Translation between non-English languages, e.g., French$\rightarrow$German, is usually done with pivoting through English, i.e., translating French (\emph{source}) input to English (\emph{pivot}) first with a French$\rightarrow$English model which is later translated to German (\emph{target}) with a English$\rightarrow$German model \cite{de2006catalan,utiyama2007comparison,wu2007pivot}.
  However, pivoting requires doubled decoding time and the translation errors are propagated or expanded via the two-step process.
  
  Therefore, it is more beneficial to build a single source$\rightarrow$target model directly for both efficiency and adequacy.
  Since non-English language pairs often have little or no parallel text, common choices to avoid pivoting in NMT are generating pivot-based synthetic data \cite{bertoldi2008phrase,chen2017teacher} or training multilingual systems \cite{firat2016zero,johnson2017google}.

  In this work, we present novel transfer learning techniques to effectively train a single, direct NMT model for a non-English language pair.
  We pre-train NMT models for source$\rightarrow$pivot and pivot$\rightarrow$target, which are transferred to a source$\rightarrow$target model.
  To optimize the usage of given source-pivot and pivot-target parallel data for the source$\rightarrow$target direction, we devise the following techniques to smooth the discrepancy between the pre-trained and final models:
  \begin{itemize}\itemsep0em
      \item Step-wise pre-training with careful parameter freezing.
      \item Additional adapter component to familiarize the pre-trained decoder with the outputs of the pre-trained encoder.
      \item Cross-lingual encoder pre-training with autoencoding of the pivot language.
  \end{itemize}
  
  Our methods are evaluated in two non-English language pairs of WMT 2019 news translation tasks: high-resource (French$\rightarrow$German) and low-resource (German$\rightarrow$Czech). We show that NMT models pre-trained with our methods are highly effective in various data conditions, when fine-tuned for source$\rightarrow$target with:
  \begin{itemize}\itemsep0em
      \item Real parallel corpus
      \item Pivot-based synthetic parallel corpus (\emph{zero-resource})
      \item None (\emph{zero-shot})
  \end{itemize}
  For each data condition, we consistently outperform strong baselines, e.g., multilingual, pivoting, or teacher-student, showing the universal effectiveness of our transfer learning schemes.
  
  The rest of the paper is organized as follows. We first review important previous works on pivot-based MT in Section \ref{sec:related}. Our three pre-training techniques are presented in Section \ref{sec:methods}. Section \ref{sec:results} shows main results of our methods with a detailed description of the experimental setups. Section \ref{sec:analysis} studies variants of our methods and reports the results without source-target parallel resources or with large synthetic parallel data. Section 6 draws conclusion of this work with future research directions.

\section{Related Work}
\label{sec:related}
In this section, we first review existing approaches to leverage a pivot language in low-resource/zero-resource MT.
They can be divided into three categories:

\begin{enumerate}\itemsep0em
    \item \label{sec:pivoting} \textbf{Pivot translation (pivoting).} The most naive approach is reusing (already trained) source$\rightarrow$pivot and pivot$\rightarrow$target models directly, decoding twice via the pivot language \cite{kauers2002interlingua,de2006catalan}.
    One can keep $N$-best hypotheses in the pivot language to reduce the prediction bias \cite{utiyama2007comparison} and improve the final translation by system combination \cite{costa2011enhancing}, which however increases the translation time even more.
    In multilingual NMT, \newcite{firat2016zero} modify the second translation step (pivot$\rightarrow$target) to use source and pivot language sentences together as the input.
    
    \item \label{sec:pivot-synth} \textbf{Pivot-based synthetic parallel data.} We may translate the pivot side of given pivot-target parallel data using a pivot$\rightarrow$source model \cite{bertoldi2008phrase}, or the other way around translating source-pivot data using a pivot$\rightarrow$target model \cite{de2006catalan}.
    For NMT, the former is extended by \newcite{zheng2017maximum} to compute the expectation over synthetic source sentences.
    The latter is also called teacher-student approach \cite{chen2017teacher}, where the pivot$\rightarrow$target model (teacher) produces target hypotheses for training the source$\rightarrow$target model (student).
    
    \item \textbf{Pivot-based model training.}
    In phrase-based MT, there have been many efforts to combine phrase/word level features of source-pivot and pivot-target into a source$\rightarrow$target system \cite{utiyama2007comparison,wu2007pivot,bakhshaei2010farsi,zahabi2013using,zhu2014improving,miura2015improving}.
    In NMT, \newcite{cheng2017joint} jointly train for three translation directions of source-pivot-target by sharing network components, where \newcite{ren2018triangular} use the expectation-maximization algorithm with the target sentence as a latent variable.
    \newcite{lu2018neural} deploy intermediate recurrent layers which are common for multiple encoders and decoders, while \newcite{johnson2017google} share all components of a single multilingual model.
    Both methods train the model for language pairs involving English but enable zero-shot translation for unseen non-English language pairs.
    For this, \newcite{ha2017effective} encode the target language as an additional embedding and filter out non-target tokens in the output.
    \newcite{lakew2017improving} combine the multilingual training with synthetic data generation to improve the zero-shot performance iteratively, where \newcite{sestorain2018zero} applies the NMT prediction score and a language model score to each synthetic example as gradient weights.
\end{enumerate}

Our work is based on transfer learning \cite{zoph2016transfer} and belongs to the third category: model training.
On the contrary to the multilingual joint training, we suggest two distinct steps: pre-training (with source-pivot and pivot-target data) and fine-tuning (with source-target data).
With our proposed methods, we prevent the model from losing its capacity to other languages while utilizing the information from related language pairs well, as shown in the experiments (Section \ref{sec:results}).

Our pivot adapter (Section \ref{sec:adapter}) shares the same motivation with the interlingua component of \newcite{lu2018neural}, but is much compact, independent of variable input length, and easy to train offline.
The adapter training algorithm is adopted from bilingual word embedding mapping \cite{xing2015normalized}.
Our cross-lingual encoder (Section \ref{sec:cross-enc}) is inspired by cross-lingual sentence embedding algorithms using NMT \cite{schwenk2017learning,schwenk2018filtering}.

Transfer learning was first introduced to NMT by \newcite{zoph2016transfer}, where only the source language is switched before/after the transfer.
\newcite{nguyen2017transfer} and \newcite{kocmi2018trivial} use shared subword vocabularies to work with more languages and help target language switches.
\newcite{kim2019effective} propose additional techniques to enable NMT transfer even without shared vocabularies.
To the best of our knowledge, we are the first to propose transfer learning strategies specialized in utilizing a pivot language, transferring a source encoder and a target decoder at the same time.
Also, for the first time, we present successful zero-shot translation results only with pivot-based NMT pre-training.

\section{Pivot-based Transfer Learning}
\label{sec:methods}

\begin{figure}[!t]
    \centering
    \includegraphics[width=\linewidth]{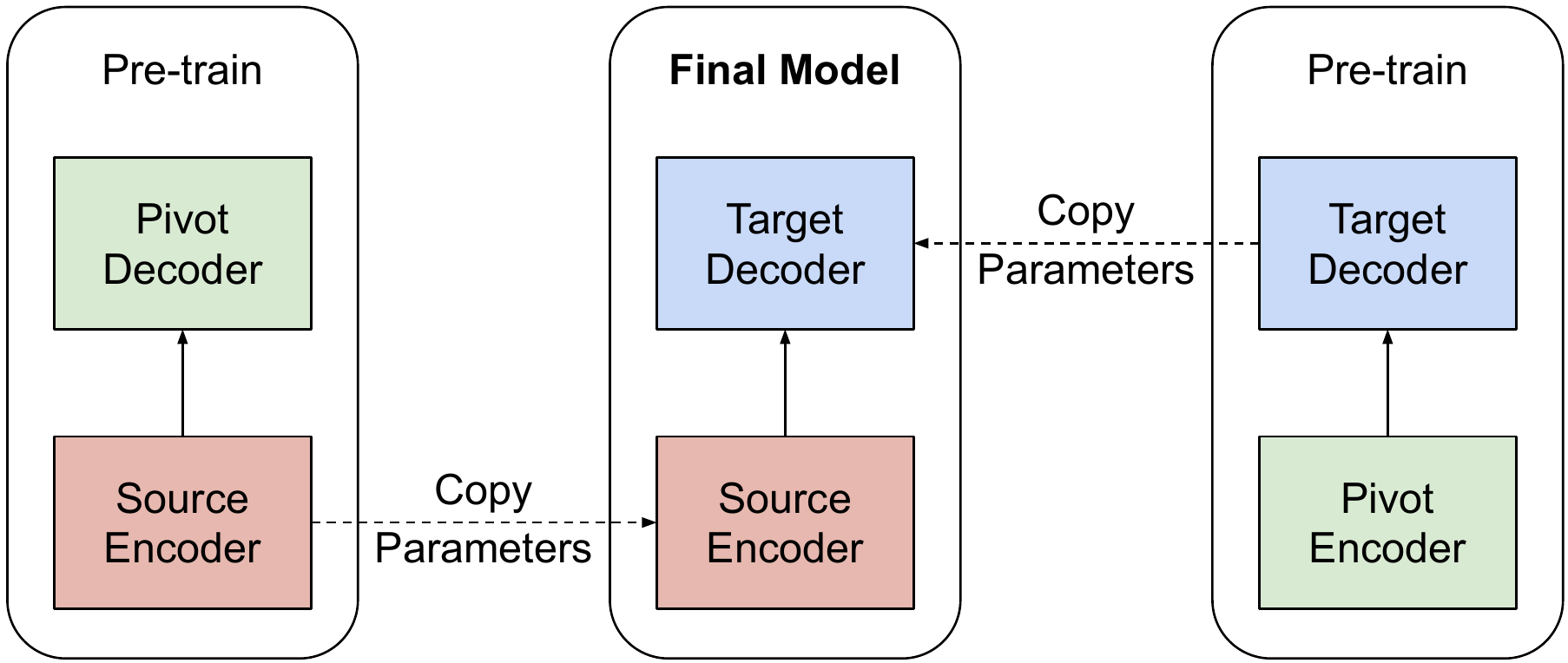}
    \caption{Plain transfer learning.}
    \label{fig:plain}
\end{figure}

Our methods are based on a simple transfer learning principle for NMT, adjusted to a usual data condition for non-English language pairs: lots of source-pivot and pivot-target parallel data, little (low-resource) or no (zero-resource) source-target parallel data.
Here are the core steps of the plain transfer (Figure \ref{fig:plain}):
\begin{enumerate}\itemsep0em
    \item Pre-train a source$\rightarrow$pivot model with a source-pivot parallel corpus and a pivot$\rightarrow$target model with a pivot-target parallel corpus.
    \item Initialize the source$\rightarrow$target model with the source encoder from the pre-trained source$\rightarrow$pivot model and the target decoder from the pre-trained pivot$\rightarrow$target model.
    \item Continue the training with a source-target parallel corpus.
\end{enumerate}
If we skip the last step (for zero-resource cases) and perform the source$\rightarrow$target translation directly, it corresponds to zero-shot translation.

Thanks to the pivot language, we can pre-train a source encoder and a target decoder without changing the model architecture or training objective for NMT. On the contrary to other NMT transfer scenarios \cite{zoph2016transfer,nguyen2017transfer,kocmi2018trivial}, this principle has no language mismatch between transferor and transferee on each source/target side.
Experimental results (Section \ref{sec:results}) also show its competitiveness despite its simplicity.

Nonetheless, the main caveat of this basic pre-training is that the source encoder is trained to be used by an English decoder, while the target decoder is trained to use the outputs of an English encoder --- not of a source encoder.
In the following, we propose three techniques to mitigate the inconsistency of source$\rightarrow$pivot and pivot$\rightarrow$target pre-training stages.
Note that these techniques are not exclusive and some of them can complement others for a better performance of the final model.

\subsection{Step-wise Pre-training}
\label{sec:step-wise}

\begin{figure}[!t]
    \centering
    \includegraphics[width=\linewidth]{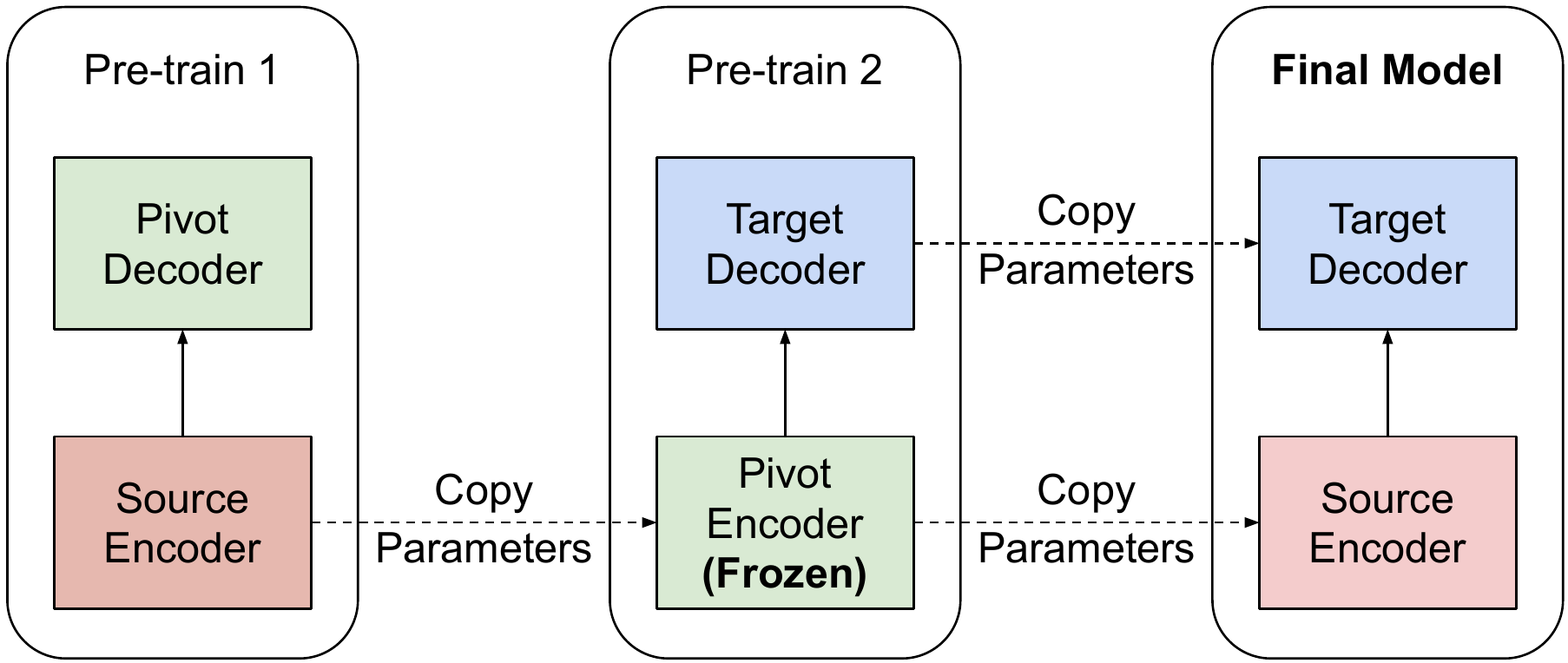}
    \caption{Step-wise pre-training.}
    \label{fig:step-wise}
\end{figure}

A simple remedy to make the pre-trained encoder and decoder refer to each other is to train a single NMT model for source$\rightarrow$pivot and pivot$\rightarrow$target in consecutive steps (Figure \ref{fig:step-wise}):
\begin{enumerate}\itemsep0em
    \item Train a source$\rightarrow$pivot model with a source-pivot parallel corpus.
    \item Continue the training with a pivot-target parallel corpus, while freezing the encoder parameters of 1.
\end{enumerate}

In the second step, a target decoder is trained to use the outputs of the pre-trained source encoder as its input.
Freezing the pre-trained encoder ensures that, even after the second step, the encoder is still modeling the source language although we train the NMT model for pivot$\rightarrow$target.
Without the freezing, the encoder completely adapts to the pivot language input and is likely to forget source language sentences.

We build a joint vocabulary of the source and pivot languages so that the encoder effectively represents both languages.
The frozen encoder is pre-trained for the source language in the first step, but also able to encode a pivot language sentence in a similar representation space.
It is more effective for linguistically similar languages where many tokens are common for both languages in the joint vocabulary.

\subsection{Pivot Adapter}
\label{sec:adapter}

\begin{figure}[!t]
    \centering
    \includegraphics[width=\linewidth]{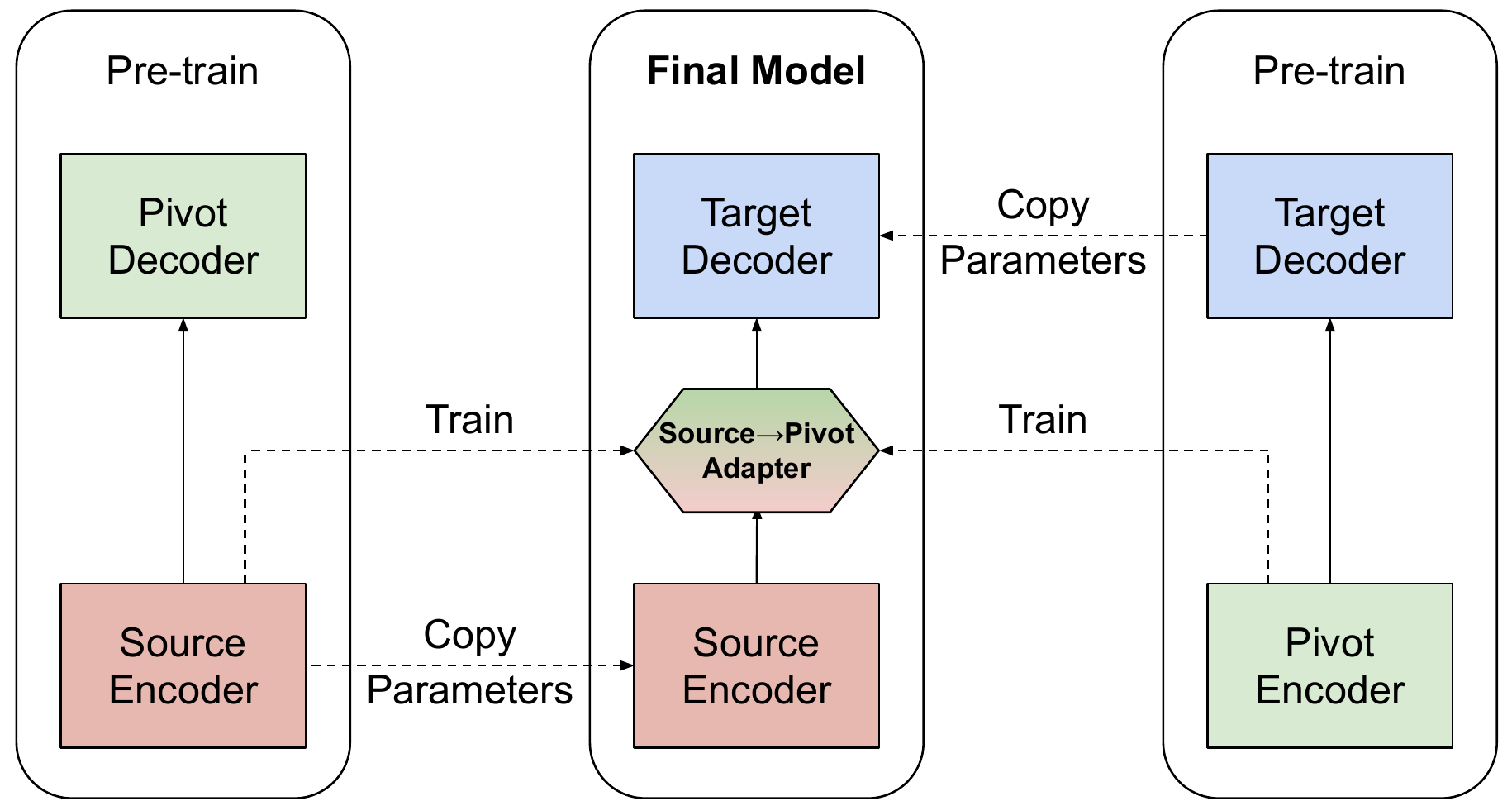}
    \caption{Pivot adapter.}
    \label{fig:adapter}
\end{figure}

Instead of the step-wise pre-training, we can also postprocess the network to enhance the connection between the source encoder and the target decoder which are pre-trained individually.
Our idea is that, after the pre-training steps, we adapt the source encoder outputs to the pivot encoder outputs to which the target decoder is more familiar (Figure \ref{fig:adapter}). We learn a linear mapping between the two representation spaces with a small source-pivot parallel corpus:
\begin{enumerate}\itemsep0em
    \item Encode the source sentences with the source encoder of the pre-trained source$\rightarrow$pivot model.
    \item Encode the pivot sentences with the pivot encoder of the pre-trained pivot$\rightarrow$target model.
    \item Apply a pooling to each sentence of 1 and 2, extracting representation vectors for each sentence pair: ($\mathbf{s}$, $\mathbf{p}$).
    \item Train a mapping $\mathbf{M}\in\mathbb{R}^{d \times d}$ to minimize the distance between the pooled representations $\mathbf{s}\in\mathbb{R}^{d \times 1}$ and $\mathbf{p}\in\mathbb{R}^{d \times 1}$, where the source representation is first fed to the mapping:
    \begin{align}
        \hat{\mathbf{M}} = \argmin_{\mathbf{M}} \sum_{\mathbf{s},\mathbf{p}} \|\mathbf{M}\mathbf{s}-\mathbf{p}\|^2
        \label{eq:min-vec}
    \end{align}
\end{enumerate}
where $d$ is the hidden layer size of the encoders. Introducing matrix notations $\mathbf{S}\in\mathbb{R}^{d \times n}$ and $\mathbf{P}\in\mathbb{R}^{d \times n}$, which concatenate the pooled representations of all $n$ sentences for each side in the source-pivot corpus, we rewrite Equation \ref{eq:min-vec} as:
\begin{align}
    \mathbf{\hat{M}} = \argmin_{\mathbf{M}} \|\mathbf{M}\mathbf{S}-\mathbf{P}\|^2
    \label{eq:min-mtx}
\end{align}
which can be easily computed by the singular value decomposition (SVD) for a closed-form solution, if we put an orthogonality constraint on $\mathbf{M}$ \cite{xing2015normalized}.
The resulting optimization is also called Procrustes problem.

The learned mapping is multiplied to encoder outputs of all positions in the final source$\rightarrow$target tuning step.
With this mapping, the source encoder emits sentence representations that lie in a similar space of the pivot encoder.
Since the target decoder is pre-trained for pivot$\rightarrow$target and accustomed to receive the pivot encoder outputs, it should process the mapped encoder outputs better than the original source encoder outputs.

\subsection{Cross-lingual Encoder}
\label{sec:cross-enc}

\begin{figure}[!t]
    \centering
    \includegraphics[width=\linewidth]{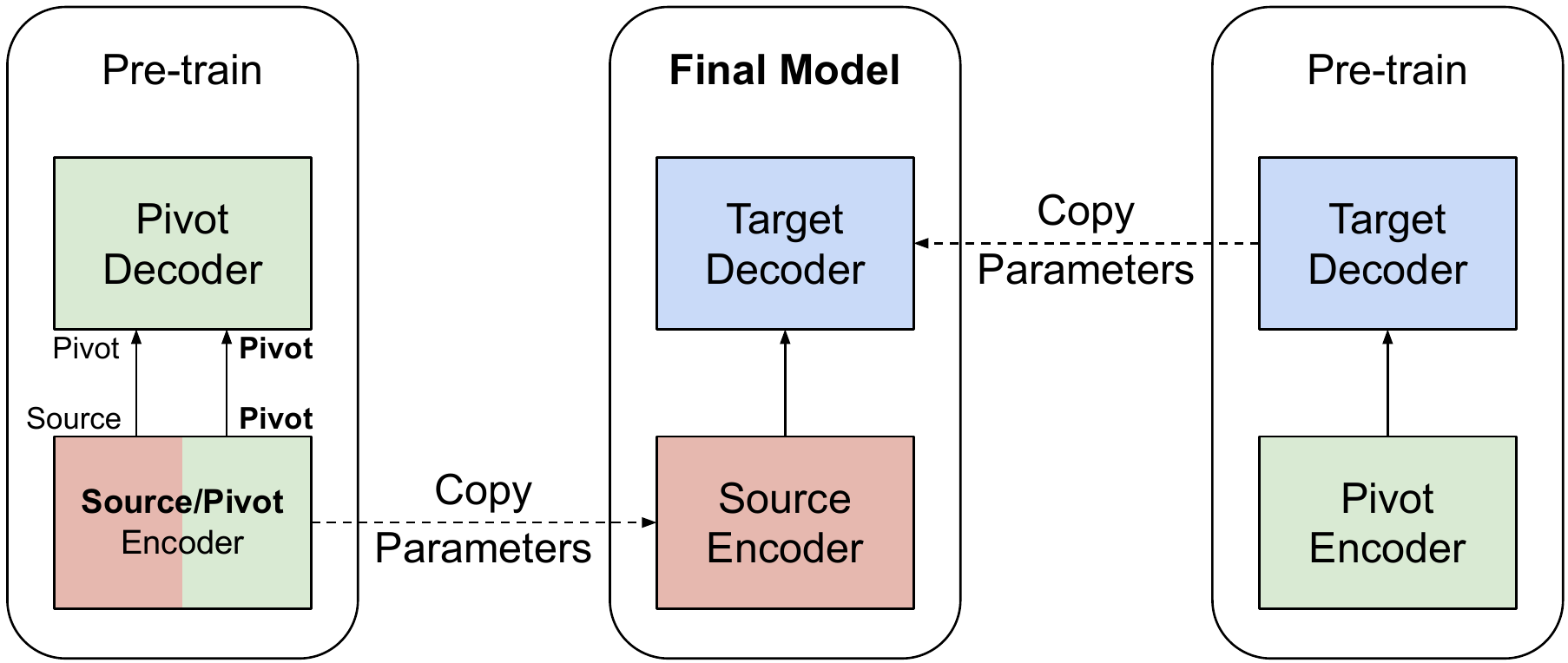}
    \caption{Cross-lingual encoder.}
    \label{fig:cross-lingual}
\end{figure}

As a third technique, we modify the source$\rightarrow$pivot pre-training procedure to force the encoder to have cross-linguality over source and pivot languages; modeling source and pivot sentences in the same mathematical space.
We achieve this by an additional autoencoding objective from a pivot sentence to the same pivot sentence (Figure \ref{fig:cross-lingual}).

The encoder is fed with sentences of both source and pivot languages, which are processed by a shared decoder that outputs only the pivot language.
In this way, the encoder is learned to produce representations in a shared space regardless of the input language, since they are used in the same decoder.
This cross-lingual space facilitates smoother learning of the final source$\rightarrow$target model, because the decoder is pre-trained to translate the pivot language.

The same input/output in autoencoding encourages, however, merely copying the input; it is said to be not proper for learning complex structure of the data domain \cite{vincent2008extracting}.
Denoising autoencoder addresses this by corrupting the input sentences by artificial noises \cite{hill2016learning}.
Learning to reconstruct clean sentences, it encodes linguistic structures of natural language sentences, e.g., word order, better than copying.
Here are the noise types we use \cite{edunov2018understanding}:
\begin{itemize}\itemsep0em
    \item Drop tokens randomly with a probability $p_\mathrm{del}$
    \item Replace tokens with a \texttt{<BLANK>} token randomly with a probability $p_\mathrm{rep}$
    \item Permute the token positions randomly so that the difference between an original index and its new index is less than or equal to $d_\mathrm{per}$
\end{itemize}
We set $p_\mathrm{del}=0.1$, $p_\mathrm{rep}=0.1$, and $d_\mathrm{per}=3$ in our experiments.\vspace{0.7em}

\noindent The key idea of all three methods is to build a closer connection between the pre-trained encoder and decoder via a pivot language.
The difference is in when we do this job:
Cross-lingual encoder (Section \ref{sec:cross-enc}) changes the encoder pre-training stage (source$\rightarrow$pivot), while step-wise pre-training (Section \ref{sec:step-wise}) modifies decoder pre-training stage (pivot$\rightarrow$target).
Pivot adapter (Section \ref{sec:adapter}) is applied after all pre-training steps. 

\section{Main Results}
\label{sec:results}

We evaluate the proposed transfer learning techniques in two non-English language pairs of WMT 2019 news translation tasks\footnote{\scriptsize{\url{http://www.statmt.org/wmt19/translation-task.html}}}: French$\rightarrow$German and German$\rightarrow$Czech.\vspace{0.7em}

\noindent\textbf{Data}\hspace{0.3cm} We used the News Commentary v14 parallel corpus and newstest2008-2010 test sets as the source-target training data for both tasks.
The newstest sets were oversampled four times.
The German$\rightarrow$Czech task was originally limited to unsupervised learning (using only monolingual corpora) in WMT 2019, but we relaxed this constraint by the available parallel data.
We used newstest2011 as a validation set and newstest2012/newstest2013 as the test sets.

Both language pairs have much abundant parallel data in source-pivot and pivot-target with English as the pivot language. Detailed corpus statistics are given in Table \ref{tab:corpus-stat}.\vspace{0.7em}

\noindent\textbf{Preprocessing}\hspace{0.3cm} We used the Moses\footnote{\scriptsize{\url{http://www.statmt.org/moses/}}} tokenizer and applied true-casing on all corpora.
For all transfer learning setups, we learned byte pair encoding (BPE) \cite{sennrich2016neural} for each language individually with 32k merge operations, except for cross-lingual encoder training with joint BPE only over source and pivot languages.
This is for modularity of pre-trained models: for example, a French$\rightarrow$English model trained with joint French/English/German BPE could be transferred smoothly to a French$\rightarrow$German model, but would not be optimal for a transfer to e.g., a French$\rightarrow$Korean model.
Once we pre-train an NMT model with separate BPE vocabularies, we can reuse it for various final language pairs without wasting unused portion of subword vocabularies (e.g., German-specific tokens in building a French$\rightarrow$Korean model).

On the contrary, baselines used joint BPE over all languages with also 32k merges.\vspace{0.7em}

\begin{table}[!t]
\centering
\begin{tabular}{cccc}
    \toprule
    & & & Words\\
    Usage & Data & Sentences & (Source)\\
    \midrule\midrule
    \multirow{2}{*}{Pre-train} & fr-en & 35M & 950M\\
    & en-de & 9.1M & 170M\\
    \midrule
    Fine-tune & fr-de & 270k & 6.9M\\
    \midrule\midrule
    \multirow{2}{*}{Pre-train} & de-en & 9.1M & 181M\\
    & en-cs & 49M & 658M\\
    \midrule
    Fine-tune & de-cs & 230k & 5.1M\\
    \bottomrule
\end{tabular}
\caption{Parallel training data statistics.}
\label{tab:corpus-stat}
\end{table}

\noindent\textbf{Model and Training}\hspace{0.3cm} The 6-layer base Transformer architecture \cite{vaswani2017attention} was used for all of our experiments.
Batch size was set to 4,096 tokens.
Each checkpoint amounts to 10k updates for pre-training and 20k updates for fine-tuning.

Each model was optimized with Adam \cite{kingma2014adam} with an initial learning rate of 0.0001, which was multiplied by 0.7 whenever perplexity on the validation set was not improved for three checkpoints. When it was not improved for eight checkpoints, we stopped the training. The NMT model training and transfer were done with the \textsc{OpenNMT} toolkit \cite{klein-etal-2017-opennmt}.

Pivot adapter was trained using the \textsc{Muse} toolkit \cite{conneau2018word}, which was originally developed for bilingual word embeddings but we adjusted for matching sentence representations.\vspace{0.7em}

\setlength{\tabcolsep}{2.5pt}
\begin{table*}[!ht]
\centering
\begin{tabular}{lcccccccccc}
    \toprule
    & & \multicolumn{4}{c}{French$\rightarrow$German} & & \multicolumn{4}{c}{German$\rightarrow$Czech} \\
    \cmidrule{3-6}\cmidrule{8-11}
    & & \multicolumn{2}{c}{\footnotesize{newstest2012}} & \multicolumn{2}{c}{\footnotesize{newstest2013}} & & \multicolumn{2}{c}{\footnotesize{newstest2012}} & \multicolumn{2}{c}{\footnotesize{newstest2013}} \\
    & & \footnotesize{\textsc{Bleu} [\%]} & \footnotesize{\textsc{Ter} [\%]} & \footnotesize{\textsc{Bleu} [\%]} & \footnotesize{\textsc{Ter} [\%]} & & \footnotesize{\textsc{Bleu} [\%]} & \footnotesize{\textsc{Ter} [\%]} & \footnotesize{\textsc{Bleu} [\%]} & \footnotesize{\textsc{Ter} [\%]}\\
    \midrule
    Direct source$\rightarrow$target & & 14.8 & 75.1 & 16.0 & 75.1 & & 11.1 & 81.1 & 12.8 & 77.7\\
    Multilingual many-to-many & & 18.7 & 71.9 & 19.5 & 72.6 & & 14.9 & 76.6 & 16.5 & 73.2\\
    Multilingual many-to-one & & 18.3 & 71.7 & 19.2 & 71.5 & & 13.1 & 79.6 & 14.6 & 75.8\\
    \midrule
    Plain transfer & & 17.5 & 72.3 & 18.7 & 71.8 & & 15.4 & 75.4 & 18.0 & 70.9\\
    \enspace + Pivot adapter & & 18.0 & 71.9 & 19.1 & 71.1 & & 15.9 & 75.0 & 18.7 & 70.3\\
    \enspace + Cross-lingual encoder & & 17.4 & 72.1 & 18.9 & 71.8 & & 15.0 & 75.9 & 17.6 & 71.4\\
    \enspace \enspace \enspace + Pivot adapter & & 17.8 & 72.3 & 19.1 & 71.5 & & 15.6 & 75.3 & 18.1 & 70.8\\
    Step-wise pre-training & & 18.6 & 70.7 & 19.9 & 70.4 & & 15.6 & 75.0 & 18.1 & 70.9\\
    \enspace + Cross-lingual encoder & & \textbf{19.5} & \textbf{69.8} & \textbf{20.7} & \textbf{69.4} & & \textbf{16.2} & \textbf{74.6} & \textbf{19.1} & \textbf{69.9}\\
    \bottomrule
\end{tabular}
\caption{Main results fine-tuned with source-target parallel data.}
\label{tab:main}
\end{table*}

\noindent\textbf{Baselines}\hspace{0.3cm} We thoroughly compare our approaches to the following baselines:
\begin{enumerate}\itemsep0em
    \item \emph{Direct source$\rightarrow$target}: A standard NMT model trained on given source$\rightarrow$target parallel data.
    \item \emph{Multilingual}: A single, shared NMT model for multiple translation directions \cite{johnson2017google}.
    \begin{itemize}\itemsep0em
        \item \emph{Many-to-many}: Trained for all possible directions among source, target, and pivot languages.
        \item \emph{Many-to-one}: Trained for only the directions \emph{to} target language, i.e., source$\rightarrow$target and pivot$\rightarrow$target, which tends to work better than many-to-many systems \cite{aharoni2019massively}.\vspace{0.3em}
    \end{itemize}
\end{enumerate}

In Table \ref{tab:main}, we report principal results after fine-tuning the pre-trained models using source-target parallel data.

As for baselines, multilingual models are better than a direct NMT model.
The many-to-many models surpass the many-to-one models; since both tasks are in a low-resource setup, the model gains a lot from related language pairs even if the target languages do not match.

Plain transfer of pre-trained encoder/decoder without additional techniques (Figure \ref{fig:plain}) shows a nice improvement over the direct baseline: up to +2.7\% \textsc{Bleu} for French$\rightarrow$German and +5.2\% \textsc{Bleu} for German$\rightarrow$Czech.
Pivot adapter provides an additional boost of maximum +0.7\% \textsc{Bleu} or -0.7\% \textsc{Ter}.

Cross-lingual encoder pre-training is proved to be not effective in the plain transfer setup.
It shows no improvements over plain transfer in French$\rightarrow$German, and 0.4\% \textsc{Bleu} worse performance in German$\rightarrow$Czech.
We conjecture that the cross-lingual encoder needs a lot more data to be fine-tuned for another decoder, where the encoder capacity is basically divided into two languages at the beginning of the fine-tuning.
On the other hand, the pivot adapter directly improves the connection to an individually pre-trained decoder, which works nicely with small fine-tuning data.

Pivot adapter gives an additional improvement on top of the cross-lingual encoder; up to +0.4\% \textsc{Bleu} in French$\rightarrow$German and +0.6\% \textsc{Bleu} in German$\rightarrow$Czech.
In this case, we extract source and pivot sentence representations from the same shared encoder for training the adapter.

Step-wise pre-training gives a big improvement up to +1.2\% \textsc{Bleu} or -1.6\% \textsc{Ter} against plain transfer in French$\rightarrow$German.
It shows the best performance in both tasks when combined with the cross-lingual encoder: up to +1.2\% \textsc{Bleu} in French$\rightarrow$German and +2.6\% \textsc{Bleu} in German$\rightarrow$Czech, compared to the multilingual baseline.
Step-wise pre-training prevents the cross-lingual encoder from degeneration, since the pivot$\rightarrow$target pre-training (Step 2 in Section \ref{sec:step-wise}) also learns the encoder-decoder connection with a large amount of data --- in addition to the source$\rightarrow$target tuning step afterwards.

Note that the pivot adapter, which inserts an extra layer between the encoder and decoder, is not appropriate after the step-wise pre-training; the decoder is already trained to correlate well with the pre-trained encoder.
We experimented with the pivot adapter on top of step-wise pre-trained models --- with or without cross-lingual encoder --- but obtained detrimental results.

Compared to pivot translation (Table \ref{tab:zero}), our best results are also clearly better in French $\rightarrow$German and comparable in German$\rightarrow$Czech.

\section{Analysis}
\label{sec:analysis}

In this section, we conduct ablation studies on the variants of our methods and see how they perform in different data conditions.

\subsection{Pivot Adapter}

\setlength{\tabcolsep}{3pt}
\begin{table}[!ht]
\centering
\begin{tabular}{ccc}
    \toprule
     & \multicolumn{2}{c}{\footnotesize{newstest2013}}\\
    Adapter Training & \footnotesize{\textsc{Bleu} [\%]} & \footnotesize{\textsc{Ter} [\%]}\\
    \midrule
    None & 18.2 & 70.7\\
    Max-pooled & 18.4 & 70.5\\
    Average-pooled & \textbf{18.7} & \textbf{70.3}\\
    \midrule \midrule
    Plain transfer & 18.0 & 70.9\\
    \bottomrule
\end{tabular}
\caption{Pivot adapter variations (German$\rightarrow$Czech). All results are tuned with source-target parallel data.}
\label{tab:adapter}
\end{table}

Firstly, we compare variants of the pivot adapter (Section \ref{sec:adapter}) in Table \ref{tab:adapter}.
The row ``None'' shows that a randomly initialized linear layer already guides the pre-trained encoder/decoder to harmonize with each other.
Of course, when we train the adapter to map source encoder outputs to pivot encoder outputs, the performance gets better.
For compressing encoder outputs over positions, average-pooling is better than max-pooling.
We observed the same trend in the other test set and in French$\rightarrow$German.

We also tested nonlinear pivot adapter, e.g., a 2-layer feedforward network with ReLU activations, but the performance was not better than just a linear adapter.

\subsection{Cross-lingual Encoder}

\setlength{\tabcolsep}{3pt}
\begin{table}[!ht]
\centering
\begin{tabular}{cccc}
    \toprule
     & & \multicolumn{2}{c}{\footnotesize{newstest2013}}\\
    Trained on & Input & \footnotesize{\textsc{Bleu} [\%]} & \footnotesize{\textsc{Ter} [\%]}\\
    \midrule
    \multirow{2}{*}{Monolingual} & Clean & 15.7 & 77.7\\
    & Noisy & 17.5 & 73.6\\
    \midrule
    \multirow{2}{*}{Pivot side of parallel} & Clean & 15.9 & 77.3\\
    & Noisy & \textbf{18.0} & \textbf{72.7}\\
    \bottomrule
\end{tabular}
\caption{Cross-lingual encoder variations (French$\rightarrow$ German). All results are in the zero-shot setting with step-wise pre-training.}
\label{tab:cross-enc}
\end{table}

Table \ref{tab:cross-enc} verifies that the noisy input in autoencoding is indeed beneficial to our cross-lingual encoder.
It improves the final translation performance by maximum +2.1\% \textsc{Bleu}, compared to using the copying autoencoding objective.

As the training data for autoencoding, we also compare between purely monolingual data and the pivot side of the source-pivot parallel data.
By the latter, one can expect a stronger signal for a joint encoder representation space, since two different inputs (in source/pivot languages) are used to produce the exactly same output sentence (in pivot language).
The results also tell that there are slight but consistent improvements by using the pivot part of the parallel data.

Again, we performed these comparisons in the other test set and German$\rightarrow$Czech, observing the same tendency in results.

\setlength{\tabcolsep}{2.5pt}
\begin{table*}[!ht]
\centering
\begin{tabular}{lcccccccccc}
    \toprule
    & & \multicolumn{4}{c}{French$\rightarrow$German} & & \multicolumn{4}{c}{German$\rightarrow$Czech} \\
    \cmidrule{3-6}\cmidrule{8-11}
    & & \multicolumn{2}{c}{\footnotesize{newstest2012}} & \multicolumn{2}{c}{\footnotesize{newstest2013}} & & \multicolumn{2}{c}{\footnotesize{newstest2012}} & \multicolumn{2}{c}{\footnotesize{newstest2013}} \\
    & & \footnotesize{\textsc{Bleu} [\%]} & \footnotesize{\textsc{Ter} [\%]} & \footnotesize{\textsc{Bleu} [\%]} & \footnotesize{\textsc{Ter} [\%]} & & \footnotesize{\textsc{Bleu} [\%]} & \footnotesize{\textsc{Ter} [\%]} & \footnotesize{\textsc{Bleu} [\%]} & \footnotesize{\textsc{Ter} [\%]}\\
    \midrule
    Multilingual many-to-many & & 14.1 & 79.1 & 14.6 & 79.1 & & 5.9 & - & 6.3 & 99.8\\
    Pivot translation & & 16.6 & 72.4 & 17.9 & 72.5 & & 16.4 & 74.5 & \textbf{19.5} & \textbf{70.1}\\
    Teacher-student & & 18.7 & 70.3 & 20.7 & 69.5 & & 16.0 & 75.0 & 18.5 & 70.9\\
    \midrule
    Plain transfer & & 0.1 & - & 0.2 & - & & 0.1 & - & 0.1 & -\\
    Step-wise pre-training & & 11.0 & 81.6 & 11.5 & 82.5 & & 6.0 & 92.1 & 6.5 & 87.8\\
    \enspace + Cross-lingual encoder & & 17.3 & 72.1 & 18.0 & 72.7 & & 14.1 & 76.8 & 16.5 & 73.5\\
    \enspace\enspace\enspace + Teacher-student & & \textbf{19.3} & \textbf{69.7} & \textbf{20.9} & \textbf{69.3} & & \textbf{16.5} & \textbf{74.6} & 19.1 & 70.2\\
    \bottomrule
\end{tabular}
\caption{Zero-resource results. Except those with the teacher-student, the results are all in the zero-shot setting, i.e., the model is not trained on any source-target parallel data. `-' indicates a \textsc{Ter} score over 100\%.}
\label{tab:zero}
\end{table*}

\subsection{Zero-resource/Zero-shot Scenarios}

If we do not have an access to any source-target parallel data (\emph{zero-resource}), non-English language pairs have two options for still building a working NMT system, given source-English and target-English parallel data:
\begin{itemize}\itemsep0em
    \item \emph{Zero-shot}: Perform source$\rightarrow$target translation using models which have not seen any source-target parallel sentences, e.g., multilingual models or pivoting (Section \ref{sec:related}.\ref{sec:pivoting}).
    \item \emph{Pivot-based synthetic data}: Generate synthetic source-target parallel data using source$\leftrightarrow$English and target$\leftrightarrow$English models (Section \ref{sec:related}.\ref{sec:pivot-synth}). Use this data to train a model for source$\rightarrow$target.
\end{itemize}

Table \ref{tab:zero} shows how our pre-trained models perform in zero-resource scenarios with the two options.
Note that, unlike Table \ref{tab:main}, the multilingual baselines exclude source$\rightarrow$target and target$\rightarrow$source directions.
First of all, plain transfer, where the encoder and the decoder are pre-trained separately, is poor in zero-shot scenarios.
It simply fails to connect different representation spaces of the pre-trained encoder and decoder.
In our experiments, neither pivot adapter nor cross-lingual encoder could enhance the zero-shot translation of plain transfer.

Step-wise pre-training solves this problem by changing the decoder pre-training to familiarize itself with representations from an already pre-trained encoder.
It achieves zero-shot performance of 11.5\% \textsc{Bleu} in French$\rightarrow$German and 6.5\% \textsc{Bleu} in German$\rightarrow$Czech (newstest2013), while showing comparable or better fine-tuned performance against plain transfer (see also Table \ref{tab:main}).

With the pre-trained cross-lingual encoder, the zero-shot performance of step-wise pre-training is superior to that of pivot translation in French$\rightarrow$German with only a single model.
It is worse than pivot translation in German$\rightarrow$Czech.
We think that the data size of pivot-target is critical in pivot translation; relatively huge data for English$\rightarrow$Czech make the pivot translation stronger.
Note again that, nevertheless, pivoting (second row) is very poor in efficiency since it performs decoding twice with the individual models.

For the second option (pivot-based synthetic data), we compare our methods against the sentence-level beam search version of the teacher-student framework \cite{chen2017teacher}, with which we generated 10M synthetic parallel sentence pairs. We also tried other variants of \newcite{chen2017teacher}, e.g., $N$-best hypotheses with weights, but there were no consistent improvements.

Due to enormous bilingual signals, the model trained with the teacher-student synthetic data outperforms pivot translation.
If tuned with the same synthetic data, our pre-trained model performs even better (last row), achieving the best zero-resource results on three of the four test sets.

We also evaluate our best German$\rightarrow$Czech zero-resource model on newstest2019 and compare it with the participants of the WMT 2019 unsupervised news translation task.
Ours yield 17.2\% \textsc{Bleu}, which is much better than the best single unsupervised system of the winner of the task (15.5\%) \cite{marie-etal-2019-nicts}.
We argue that, if one has enough source-English and English-target parallel data for a non-English language pair, it is more encouraged to adopt pivot-based transfer learning than unsupervised MT --- even if there is no source-target parallel data.
In this case, unsupervised MT unnecessarily restricts the data condition to using only monolingual data and its high computational cost does not pay off; simple pivot-based pre-training steps are more efficient and effective.

\setlength{\tabcolsep}{2.5pt}
\begin{table*}[!ht]
\centering
\begin{tabular}{lcccccccccc}
    \toprule
    & & \multicolumn{4}{c}{French$\rightarrow$German} & & \multicolumn{4}{c}{German$\rightarrow$Czech} \\
    \cmidrule{3-6}\cmidrule{8-11}
    & & \multicolumn{2}{c}{\footnotesize{newstest2012}} & \multicolumn{2}{c}{\footnotesize{newstest2013}} & & \multicolumn{2}{c}{\footnotesize{newstest2012}} & \multicolumn{2}{c}{\footnotesize{newstest2013}} \\
    & & \footnotesize{\textsc{Bleu} [\%]} & \footnotesize{\textsc{Ter} [\%]} & \footnotesize{\textsc{Bleu} [\%]} & \footnotesize{\textsc{Ter} [\%]} & & \footnotesize{\textsc{Bleu} [\%]} & \footnotesize{\textsc{Ter} [\%]} & \footnotesize{\textsc{Bleu} [\%]} & \footnotesize{\textsc{Ter} [\%]}\\
    \midrule
    Direct source$\rightarrow$target & & 20.1 & 69.8 & 22.3 & 68.7 & & 11.1 & 81.1 & 12.8 & 77.7\\
    \enspace + Synthetic data & & 21.1 & 68.2 & 22.6 & 68.1 & & 15.7 & 76.5 & 18.5 & 72.0\\
    \midrule
    Plain transfer & & 21.8 & 67.6 & 23.1 & 67.5 & & 17.6 & 73.2 & 20.3 & 68.7\\
    \enspace + Pivot adapter & & 21.8 & 67.6 & 23.1 & 67.6 & & \textbf{17.6} & \textbf{73.0} & \textbf{20.9} & \textbf{68.3}\\
    \enspace + Cross-lingual encoder & & 21.9 & 67.7 & 23.4 & 67.4 & & 17.5 & 73.5 & 20.3 & 68.7\\
    \enspace \enspace \enspace + Pivot adapter & & \textbf{22.1} & \textbf{67.5} & 23.3 & 67.5 & & 17.5 & 73.2 & 20.6 & 68.5\\
    Step-wise pre-training & & 21.8 & 67.8 & 23.0 & 67.8 & & 17.3 & 73.6 & 20.0 & 69.2\\
    \enspace + Cross-lingual encoder & & 21.9 & 67.6 & \textbf{23.4} & \textbf{67.4} & & 17.5 & 73.1 & 20.5 & 68.6\\
    \bottomrule
\end{tabular}
\caption{Results fine-tuned with a combination of source-target parallel data and large synthetic data. French$\rightarrow$German task used larger real parallel data than Table \ref{tab:main}.}
\label{tab:large}
\end{table*}

\subsection{Large-scale Results}

We also study the effect of pivot-based transfer learning in more data-rich scenarios: 1) with large synthetic source-target data (German$\rightarrow$Czech), and 2) with larger real source-target data in combination with the synthetic data (French$\rightarrow$German).
We generated synthetic parallel data using pivot-based back-translation \cite{bertoldi2008phrase}: 5M sentence pairs for German$\rightarrow$Czech and 9.1M sentence pairs for French$\rightarrow$German.
For the second scenario, we also prepared 2.3M more lines of French$\rightarrow$German real parallel data from Europarl v7 and Common Crawl corpora.

Table \ref{tab:large} shows our transfer learning results fine-tuned with a combination of given parallel data and generated synthetic parallel data.
The real source-target parallel data are oversampled to make the ratio of real and synthetic data to be 1:2.
As expected, the direct source$\rightarrow$target model can be improved considerably by training with large synthetic data.

Plain pivot-based transfer outperforms the synthetic data baseline by up to +1.9\% \textsc{Bleu} or -3.3\% \textsc{Ter}.
However, the pivot adapter or cross-lingual encoder gives marginal or inconsistent improvements over the plain transfer.
We suppose that the entire model can be tuned sufficiently well without additional adapter layers or a well-curated training process, once we have a large source-target parallel corpus for fine-tuning.

\section{Conclusion}

In this paper, we propose three effective techniques for transfer learning using pivot-based parallel data.
The principle is to pre-train NMT models with source-pivot and pivot-target parallel data and transfer the source encoder and the target decoder. To resolve the input/output discrepancy of the pre-trained encoder and decoder, we 1) consecutively pre-train the model for source$\rightarrow$pivot and pivot$\rightarrow$target, 2) append an additional layer after the source encoder which adapts the encoder output to the pivot language space, or 3) train a cross-lingual encoder over source and pivot languages.

Our methods are suitable for most of the non-English language pairs with lots of parallel data involving English. Experiments in WMT 2019 French$\rightarrow$German and German$\rightarrow$Czech tasks show that our methods significantly improve the final source$\rightarrow$target translation performance, outperforming multilingual models by up to +2.6\% \textsc{Bleu}. The methods are applicable also to zero-resource language pairs, showing a strong performance in the zero-shot setting or with pivot-based synthetic data. We claim that our methods expand the advances in NMT to many more non-English language pairs that are not yet studied well.

Future work will be zero-shot translation without step-wise pre-training, i.e., combining individually pre-trained encoders and decoders freely for a fast development of NMT systems for a new non-English language pair.

\section*{Acknowledgments}

\begin{center}
\vspace{0.5em}
\includegraphics[width=0.25\textwidth]{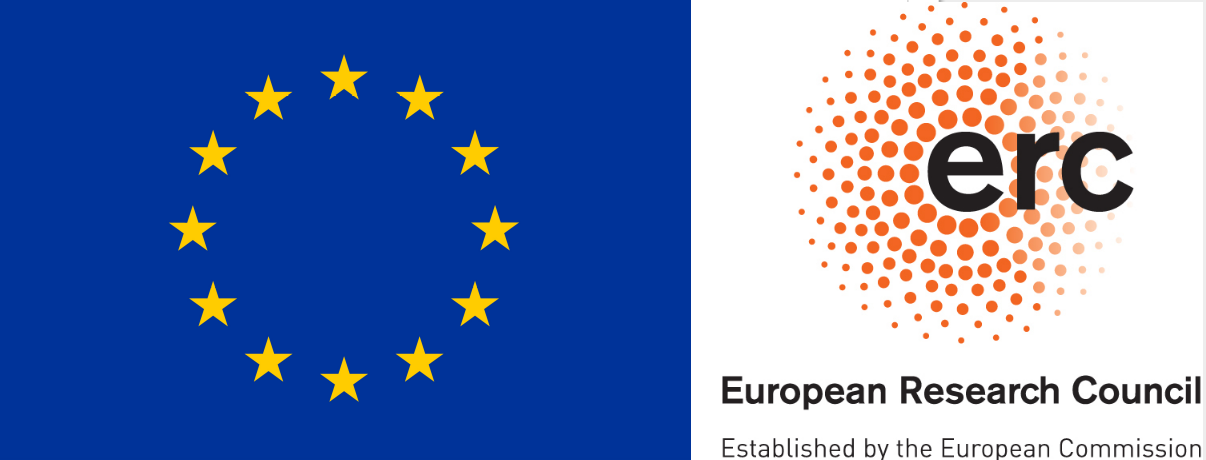}
\hspace{6pt}
\includegraphics[width=0.2\textwidth]{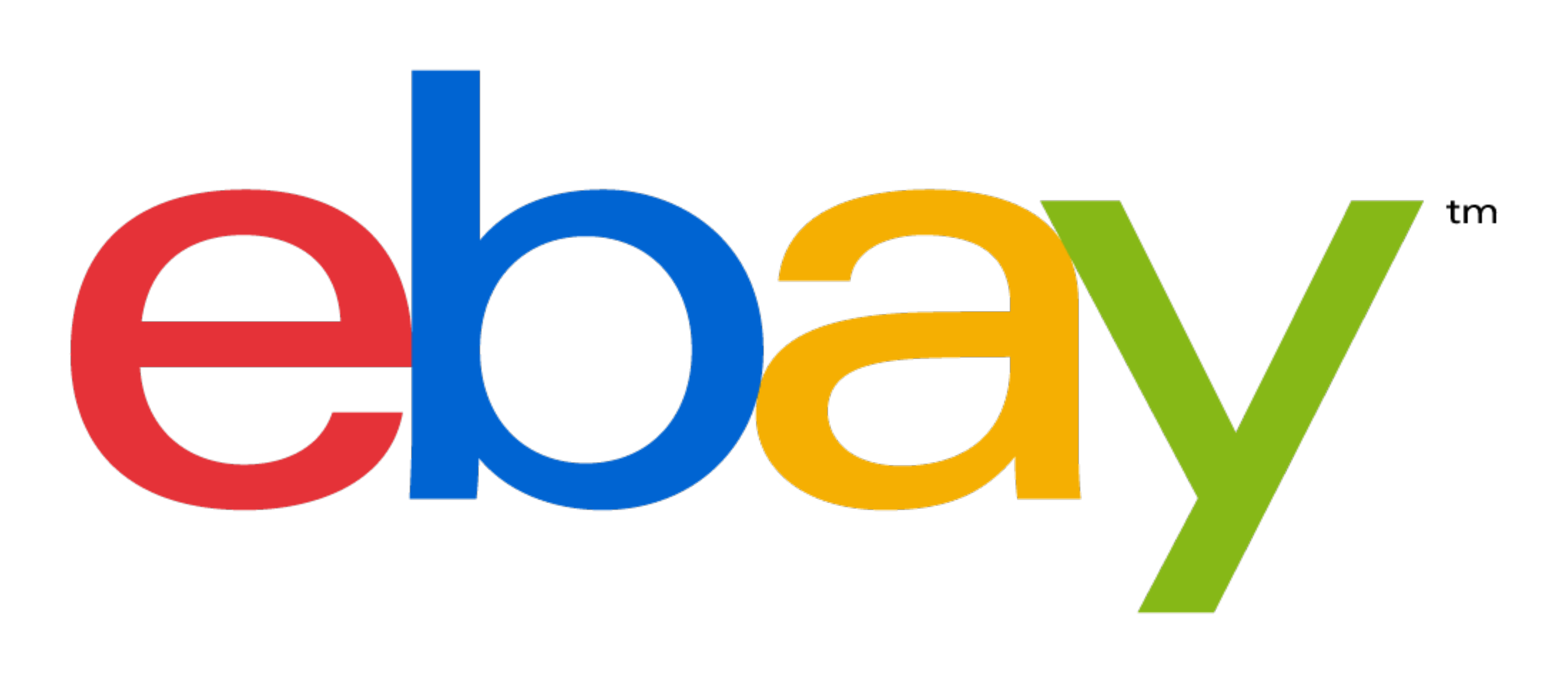}
\end{center}
\vspace{0.5em}
This work has received funding from the European Research Council (ERC) (under the European Union's Horizon 2020 research and innovation programme, grant agreement No 694537, project "SEQCLAS") and eBay Inc. The work reflects only the authors' views and none of the funding agencies is responsible for any use that may be made of the information it contains.

\bibliographystyle{acl_natbib}
\bibliography{references}

\end{document}